\documentclass[smallextended]{svjour3}        \usepackage{graphicx}
\usepackage{fancyvrb}
\usepackage[utf8]{inputenc}
\usepackage{algorithm}
\usepackage[noend]{algorithmic}

\journalname{Arxiv paper}
\begin{document}
\title{\begin{center}A Tale of Two Toolkits.\\ Report the Second: Bake Off Redux. \end{center}Chapter 1. Dictionary Based Classifiers}

\titlerunning{Bake off redux: dictionary based}        
\author{Anthony Bagnall and James Large and Matthew Middlehurst}
\institute{
Anthony Bagnall, ajb@uea.ac.uk, https://orcid.org/0000-0003-2360-8994 \\
James Large, james.large@uea.ac.uk, https://orcid.org/0000-0002-2357-3798 \\
Matthew Middlehurst, m.middlehurst@uea.ac.uk, https://orcid.org/0000-0002-3293-8779 \\
School of Computing Sciences, University of East Anglia
}
\maketitle

\begin{abstract}

Time series classification (TSC) is the problem of learning labels from time dependent data. One class of algorithms is derived from a bag of words approach. A window is run along a series, the subseries is shortened and discretised to form a word, then features are formed from the histogram of frequency of occurrence of words. We call this type of approach to TSC dictionary based classification. We compare four dictionary based algorithms in the context of a wider project to update the great time series classification bakeoff, a comparative study published in 2017. We experimentally characterise the algorithms in terms of predictive performance, time complexity and space complexity. We find that we can improve on the previous best in terms of accuracy, but this comes at the cost of time and space. Alternatively, the same performance can be achieved with far less cost. We review the relative merits of the four algorithms before suggesting a path to possible improvement.


\keywords{time series classification, sktime, tsml, bake off, bag of words}
\end{abstract}

\section{Introduction}
An experimental comparison of algorithms for time series classification conducted in 2015, referred to as a bake off~\cite{bagnall17bakeoff}, compared over 20 time series classification algorithms, most of which were proposed in the period 2010-2015, on the 85 problems that were in the UCR time series classification archive~\cite{dau18archive} at that time. Officially published in 2017, the bake off paper has, we believe, had an impact on the field and encouraged more people to work in the area. Since the bake off, the field has advanced. Numerous algorithms have been proposed that may offer greater accuracy for a sub class of problems, or provide orders of magnitude speed up over existing techniques. Software tools are in development that make comparison and reproducability much easier. Deep learning has seen a large uptake in use for TSC, and many believe it will supersede other approaches, as has happened in fields such as vision and speech.

The UCR time series repository has been expanded to include more univariate problems~\cite{dau18archive}. Hence, we feel it worthwhile repeating the bake off exercise with new algorithms and data, including a more in depth analysis of the strengths and weaknesses of alternative approaches. We conduct the experiments with two new toolkits that include time series classification functionality. sktime\footnote{https://github.com/alan-turing-institute/sktime} is an open source, Python based, sklearn compatible toolkit for time series analysis. sktime is designed to provide a unifying API for a range of time series tasks such as annotation, prediction and forecasting (see~\cite{loning19sktime} for a description of the overarching design of sktime and~\cite{bagnall19toolkits} for an experimental comparison of some of the classification algorithms available). The Java toolkit for time series machine learning, tsml\footnote{https://github.com/uea-machine-learning/tsml}, is Weka compatible and is the descendent of the codebase used to perform the bake off. We are using these two toolkits to perform the new bake off. This evaluation will be more extensive than previous work and not only focus on predictive performance: we also experimentally review run time and memory requirements. This is a large piece of work which we will split into separate arxiv papers (or chapters), which will be summarised in a final paper. The chapters will be split based on the taxonomy proposed in the bake off, which is based on the core nature of discriminatory features that are used be an algorithm.

\textbf{Dictionary based} approaches, i.e. algorithms that adapt the bag of words approach commonly used in computer vision are the subject of this work. Other categories of algorithm are based on: distance; shapelets; frequency/spectral; intervals; deep learning; and hybrids. Dictionary based classifiers work by sliding a window (size $w$) along each series, converting each window into a discrete word (length $l$) from a fixed alphabet of $\alpha$ symbols, forming a dictionary of all words and a resulting histogram of frequency counts, then classifying based on word histograms or distributions. We welcome collaboration, and if there are published dictionary approaches of which we are unaware, we will be happy to work with you to include the algorithm in either toolkit.

\section{Data}
\label{sec:data}
The UCR archive was expanded to 128 problems in 2018~\cite{dau18archive}. Of the extra 43 datasets, 15 are either of unequal length or contain missing values. These datasets are available from the UCR webpage\footnote{https://www.cs.ucr.edu/$\sim$eamonn/time\_series\_data\_2018/} or from timeseriesclassification.com. The new equal length series are summarised in Table 1.

\begin{figure}[!ht]
	\centering
\begin{tabular}{cc}
	
       \includegraphics[width =6cm,trim={2cm 4cm 2cm 4cm},clip]{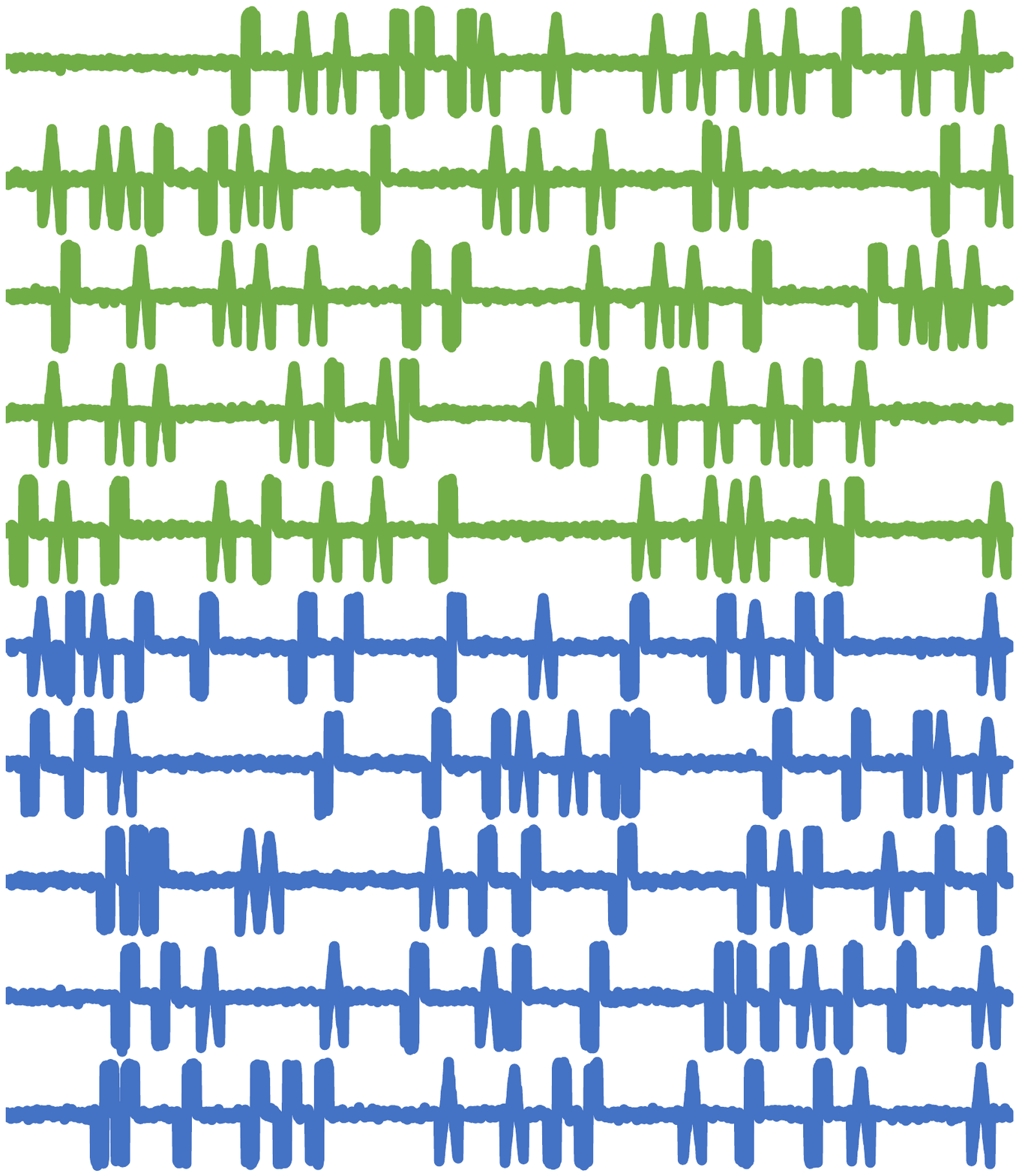}         &     	

       \includegraphics[width =6cm,trim={2cm 4cm 2cm 4cm},clip]{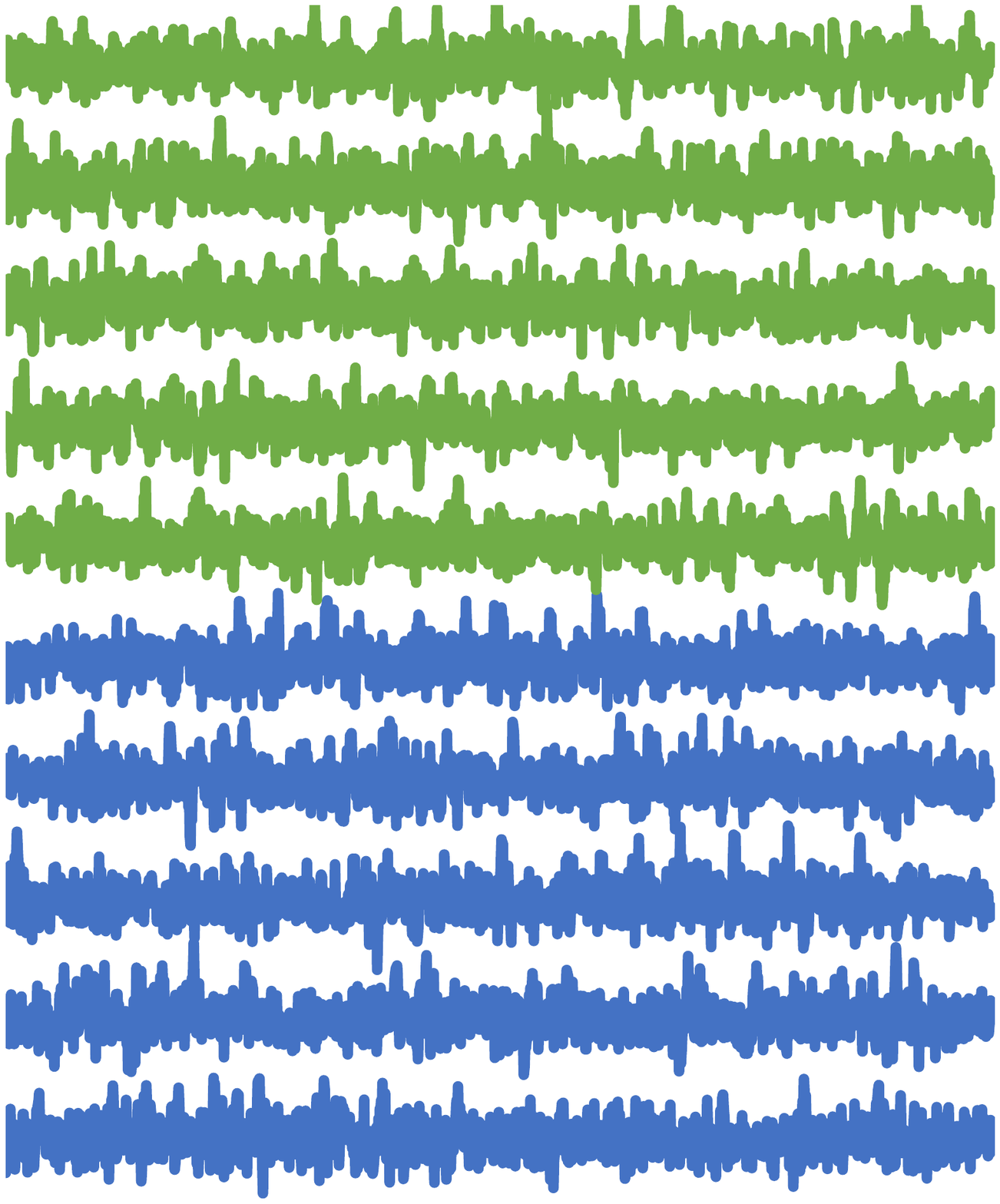}

       \end{tabular}
       \caption{Examples of simulated dictionary data for a two class problem. Class is defined by colour, the top five cases are of one class, the bottom five of another. Both classes contain examples of two distinct shapes, but each shape occurs more commonly in one class than the other. The first class contains more spike shapes than step shapes. The location of shapes is independent and randomly selected for each case, although we do not allow shapes to overlap.}.
       \label{fig:simulation}
\end{figure}

\begin{table}[h]
\label{tab:data}
\caption{The 27 new UCR data sets that have no missing values and are all of equal length.}
\begin{tabular}{|l|l|l|l|l|l|}
\hline
Name                     & Train Size & Test Size & \#Classes & Length & Type         \\
\hline
ACSF1                    & 100        & 100       & 10           & 1460   & Device       \\
BME                      & 30         & 150       & 3            & 128    & Simulated    \\
Chinatown                & 20         & 345       & 2            & 24     & Traffic      \\
Crop                     & 7200       & 16800     & 24           & 46     & Image        \\
EOGHorizontalSignal      & 362        & 362       & 12           & 1250   & EOG          \\
EOGVerticalSignal        & 362        & 362       & 12           & 1250   & EOG          \\
EthanolLevel             & 504        & 500       & 4            & 1751   & Spectro      \\
FreezerRegularTrain      & 150        & 2850      & 2            & 301    & Sensor       \\
FreezerSmallTrain        & 28         & 2850      & 2            & 301    & Sensor       \\
GunPointAgeSpan          & 135        & 316       & 2            & 150    & Motion       \\
GunPointMaleVersusFemale & 135        & 316       & 2            & 150    & Motion       \\
GunPointOldVersusYoung   & 135        & 316       & 2            & 150    & Motion       \\
HouseTwenty              & 34         & 101       & 2            & 3000   & Device       \\
InsectEPGRegularTrain    & 62         & 249       & 3            & 601    & EPG          \\
InsectEPGSmallTrain      & 17         & 249       & 3            & 601    & EPG          \\
MixedShapes              & 500        & 2425      & 5            & 1024   & Image        \\
MixedShapesSmallTrain    & 100        & 2425      & 5            & 1024   & Image        \\
PigAirwayPressure        & 104        & 208       & 52           & 2000   & Hemodynamics \\
PigArtPressure           & 104        & 208       & 52           & 2000   & Hemodynamics \\
PigCVP                   & 104        & 208       & 52           & 2000   & Hemodynamics \\
PowerCons                & 180        & 180       & 2            & 144    & Power        \\
Rock                     & 20         & 50        & 4            & 2844   & Spectrum     \\
SemgHandGenderCh2        & 300        & 600       & 2            & 1500   & Spectrum     \\
SemgHandMovementCh2      & 450        & 450       & 6            & 1500   & Spectrum     \\
SemgHandSubjectCh2       & 450        & 450       & 5            & 1500   & Spectrum     \\
SmoothSubspace           & 150        & 150       & 3            & 15     & Simulated    \\
UMD                      & 36         & 144       & 3            & 150    & Simulated   \\
\hline
\end{tabular}
\end{table}
Data simulations for time series classification were used in research conducted after the bake off~\cite{lines18hive}. A simulator that creates data idealised for each algorithm class is provided. These are described in detail in~\cite{bagnall17simulators}. We use these simulators to explore the time and memory characteristics of the classifiers in a controllable manner. Figure~\ref{fig:simulation} gives two examples of simulated dictionary data, the first with low noise for visualisation purposes, the second with standard white noise. The key discriminatory feature for this data is the number of repeating shapes in each class. Intuitively, dictionary based classifiers should be best for this type of data. We can control the complexity of the learning task by varying the number of occurrences of shapes in each class and the level of noise. The computational and memory requirements can be assessed by varying the length of series and the number of training cases.

\section{Dictionary Based Classifiers}
\label{sec:dictionarybased}

The bake off found that Bags of Symbolic-Fourier-Approximation Symbols (BOSS)~\cite{schafer15boss} was the most accurate dictionary based classifier. Hence, it forms our benchmark for new dictionary based approaches. BOSS is described in detail in Section~\ref{sec:boss}. A deconstruction of the effect of the components of dictionary based classifiers~\cite{large19dictionary} lead to an extension of BOSS based on spatial pyramids, called S-BOSS, described in Section~\ref{sec:sboss}. One of the problems with BOSS is that it can be memory and time inefficient. cBOSS (Section~\ref{sec:cboss}) addresses the scalability issues of BOSS~\cite{middlehurst19scalable} by altering the ensemble structure. WEASEL~\cite{schafer17fast} is a dictionary based classifier by the same team that produced BOSS. It is based on feature selection from histograms for a linear model (see Section~\ref{sec:weasel}).

The only other candidate we know of for possible inclusion is Mr-SEQL~\cite{nguyen19interpretable}. Mr-SEQL uses words from sliding windows and discretisation in a dictionary-like way, but it is designed to classify based on the presence or absence of words rather than their frequency. Thus it more properly belongs in the Shapelet camp and shall appear in that chapter.


\subsection{Bags of Symbolic-Fourier-Approximation Symbols (BOSS)~\cite{schafer15boss}}
\label{sec:boss}

BOSS has the following defining characteristics. It converts to words through first finding the Fourier transform, then discretising the first $l$ Fourier terms into $\alpha$ symbols to form a word, using a bespoke supervised discretisation algorithm called Multiple Coefficient Binning (MCB). It uses a non-symmetric distance function in conjunction with a nearest neighbour classifier. The final classifier is an ensemble of individual BOSS classifiers found through first fitting and evaluating a large number of individual classifiers, then retaining only those within 92\% of the best classifier. Algorithm~\ref{alg:boss} gives a more formal description of the bag forming process of an individual BOSS classifier.
        \begin{algorithm}[h]
        	\caption{baseBOSS(A list of $n$ time series of length $m$, ${\bf T}=({\bf X,y})$)}
        	\label{boss}
        	\begin{algorithmic}[1]
        \REQUIRE the word length $l$, the alphabet size $\alpha$, the window length $w$, normalisation parameter $p$
        		\STATE Let ${\bf H}$ be a list of $n$ histograms $({\bf h}_1,\ldots,{\bf h}_n)$
        		\STATE Let ${\bf B}$ be a matrix of $l$ by $\alpha$ breakpoints found by MCB
        		\FOR {$i \leftarrow  1$ to $n$}
        			\FOR {$j \leftarrow 1$ to $m-w+1$}
        				\STATE ${\bf s}\leftarrow x_{i,j} \ldots x_{i,j+w-1}$
        				\IF{$p$}
        				    \STATE $s \leftarrow $normalise($s$)
        				\ENDIF
        				\STATE ${\bf q} \leftarrow$ DFT(${\bf s}, l, \alpha$,$p$) \COMMENT{ {\em {\bf q} is a vector of the complex DFT coefficients}}
        				\IF{$p$}
            			    \STATE ${\bf q'} \leftarrow (q_2 \ldots q_{l/2+1})$
            			\ELSE
            			    \STATE ${\bf q'} \leftarrow (q_1 \ldots q_{l/2})$
            			\ENDIF
        				\STATE ${\bf r} \leftarrow$ SFAlookup(${\bf q', B}$)
        				\IF{${\bf r} \neq {\bf p}$}
        					\STATE $pos \leftarrow $index(${\bf r}$)
        					\STATE ${h}_{i,pos} \leftarrow {h}_{i,pos} + 1$
        				\ENDIF
        				\STATE ${\bf p} \leftarrow {\bf r} $
        			\ENDFOR
        		\ENDFOR
        	\end{algorithmic}
        	\label{alg:boss}
        \end{algorithm}

The BOSS ensemble (also referred to as just BOSS), evaluates all BOSS base classifiers in the range $ w \in \{10 \ldots m\}$ with $m/4$ values where m is the length of the series, $l \in \{16, 14, 12, 10, 8\}$ and $p \in \{true,false\}$. $\alpha$ stays at the default value of 4.


\subsection{Contract BOSS (cBOSS)~\cite{middlehurst19scalable}}
\label{sec:cboss}

Due to its grid-search and method of retaining ensemble members BOSS is unpredictable in its time and memory resource usage, and has been found to be sluggish or outright infeasible to build for larger problems. cBOSS~\cite{middlehurst19scalable} significantly speeds up BOSS while retaining accuracy and introduces contracting functionality by improving how BOSS forms its ensemble of base BOSS classifiers. cBOSS uses the same parameter space as BOSS, but utilises a filtered random selection of its ensemble members. A new parameter $k$ for the number of parameter samples is introduced, of which the top $s$ with the highest accuracy are kept for the final ensemble. The $k$ parameter is replaceable with a time limit $t$ through contracting. An exponential weighting scheme is introduced, setting each ensembles members weight as its train accuracy to the power of 4. Each ensemble member is built on a subsample of the train data, using a randomly selected 70\% from the whole training set. cBOSS was shown to be an order of magnitude faster than BOSS on both small and large datasets from the UCR archive while showing no significant difference in accuracy.

Algorithm~\ref{alg:cBOSS} describes the decision procedure for search and maintaining individual BOSS classifiers for cBOSS.

        \begin{algorithm}[h]
        	\caption{cBOSS(A list of $n$ cases length $m$, ${\bf T}=({\bf X,y})$)}
        	\label{alg:cBOSS}
        	\begin{algorithmic}[1]
        \REQUIRE the number of parameter samples $k$, the max ensemble size $s$
\STATE Let $w$ be window length, $l$ be word length, $p$ be normalise/not normalise and $\alpha$ be  alphabet size.
\STATE Let ${\bf C}$ be a list of $s$ BOSS classifiers $({\bf c}_1,\ldots,{\bf c}_s)$
        		\STATE Let ${\bf E}$ be a list of $s$ classifier weights $({\bf e}_1,\ldots,{\bf e}_s)$
        		\STATE Let ${\bf R}$ be a set of possible BOSS parameter combinations
        		\STATE $i \leftarrow 0$
        		\STATE $lowest\_acc \leftarrow \infty, lowest\_acc\_idx \leftarrow \infty$
        		\WHILE {$i < k$ AND $|{\bf R}| > 0$}
        		    \STATE $[l,a,w,p] \leftarrow random\_sample({\bf R}) $
        		    \STATE $ {\bf R} = {\bf R} \setminus\{[l,a,w,p]\} $
        		
        		    \STATE ${\bf T'} \leftarrow$ subsample\_data(${\bf T}$)
        		    \STATE $cls \leftarrow$ baseBOSS(${\bf T'},l,a,w,p$)
        		    \STATE $acc \leftarrow$ LOOCV($cls$) \COMMENT{ {\em train data accuracy}}
        		    \IF{$i < s$}
        		        \IF{$acc < lowest\_acc$}
        		            \STATE $lowest\_acc \leftarrow acc$, $lowest\_acc\_idx \leftarrow i$
        		        \ENDIF
        		        \STATE $c_i \leftarrow cls$, $e_i \leftarrow acc^4$
        		    \ELSIF{$acc > lowest\_acc$}
        		        \STATE $c_{lowest\_acc\_idx} \leftarrow cls$, $e_{lowest\_acc\_idx} \leftarrow acc^4$
        		        \STATE $[lowest\_acc,lowest\_acc\_idx] \leftarrow$ find\_new\_lowest\_acc(${\bf C}$)
        		    \ENDIF
        		    \STATE $i \leftarrow i+1$
        		\ENDWHILE
        	\end{algorithmic}
        \end{algorithm}

\subsection{BOSS with Spatial Pyramids (S-BOSS)~\cite{large19dictionary}}
\label{sec:sboss}

BOSS intentionally ignores the locations of words in series. For some datasets we know that the locations of certain discriminatory subsequences are important, however. Some words may gain importance only when in a particular location, or a mutually occurring word may be indicative of different classes depending on when it occurs. Spatial pyramids~\cite{lazebnik06pyramid} bring some temporal information back into the bag-of-words paradigm, and have been successfully used to improve it in computer vision problems~\cite{uijlings13selective}. While a standard bag-of-words approach creates a single bag to represent the entire object globally, a spatial pyramid recursively divides the space (into halves for 1D series, quadrants for 2D images, etc.) and builds a bag on each divided section at each scale. The height of the pyramid defines how many times the instance is divided, and therefore how fine the locational information can be. Once a bag for each section at each level is created, these are weighted to give more local bags higher importance than more global bags. These are concatenated to form an elongated feature vector per instance, proportionally longer than the original bag dependent on the height of the pyramid. The weighting means that mutually occurring words in same area of the series contribute more to the similarity between two series than mutually occurring words on a global scale.

S-BOSS incorporates the spatial pyramids technique into the BOSS algorithm. S-BOSS creates BOSS transforms using the standard procedure, forming what can be seen as S-BOSS transforms with a singe level in the pyramid - the global level. An additional degree of optimisation is then performed to find the best pyramid height $h \in \{1,2,3\}$. The height defines the importance of localisation for this transform. In terms of computation time, this is more efficient than it may initially sound since no extra transformative work is required to build the elongated bags if the positions of words are saved in the original transform. The elongated feature vectors do, however, affect the time and memory-requirements of the 1NN classifier that is used to evaluate each possible height and finally classify new cases.

The additional pyramid height searching work over the standard BOSS procedure is described in Algorithm~\ref{alg:S-BOSS}. The new work for S-BOSS is defined by lines 10 to 15, otherwise the procedure to build the ensemble is the same. For a more in-depth explanation of \textit{divide\_and\_concatenate\_bags(cls)}, see~\cite{large19dictionary}.


        \begin{algorithm}[h]
        	\caption{S-BOSS(A list of $n$ cases length $m$, ${\bf T}=({\bf X,y})$)}
        	\label{alg:S-BOSS}
        	\begin{algorithmic}[1]
                \REQUIRE the set of possible $[a,w,p]$ parameter combinations ${\bf R}$, the set of possible $[l]$ parameter values ${\bf L}$, the maximum pyramid height $H$

                \STATE Let ${\bf C}$ be a list of $s$ BOSS classifiers $({\bf c}_1,\ldots,{\bf c}_s)$
        		\FOR {$i \leftarrow  1$ to $|{\bf L}|$}
        		    \STATE $bestAcc \leftarrow 0, bestCls \leftarrow \O$
            		\FOR {$j \leftarrow  1$ to $|{\bf R}|$}
            		    \STATE $[a,w,p] \leftarrow {\bf R_j} $
            		    \STATE $cls \leftarrow$ baseBOSS(${\bf T},L_i,a,w,p$)
            		    \STATE $acc \leftarrow$ LOOCV($cls$) \COMMENT{{\em train data accuracy}}
            		    \IF{$acc > bestAcc$}
        		            \STATE $bestAcc \leftarrow acc, bestCls \leftarrow cls$
        		        \ENDIF
            		\ENDFOR
            		\STATE $cls \leftarrow bestCls$
            		\FOR {$h \leftarrow  1$ to $H$}
            		    \STATE $cls \leftarrow divide\_and\_concatenate\_bags(cls)$
            		    \STATE $acc \leftarrow$ LOOCV($cls$) \COMMENT{{\em train data accuracy}}
            		    \IF{$acc > bestAcc$}
        		            \STATE $bestAcc \leftarrow acc, bestCls \leftarrow cls$
        		        \ENDIF
            		\ENDFOR
            		
            		\STATE ${\bf C}_i \leftarrow bestCls$
        		\ENDFOR
        		\STATE $keep\_within\_best({\bf C}, 0.92)$ \COMMENT{{\em keep those cls with train accuracy within 0.92 of the best}}
       	\end{algorithmic}
        \end{algorithm}

\subsection{Word Extraction for Time Series Classification (WEASEL)~\cite{schafer17fast}}
\label{sec:weasel}

Like BOSS, WEASEL performs a Fourier transform on each window, creates words by discretisation, and forms histograms of words counts. It also does this for a range of window sizes and word lengths. However, there are important differences. WEASEL is not an ensemble NN classifiers. Instead, WEASEL constructs a single feature space from concatenated histograms for different parameter values, then uses logistic regression and feature selection. Histograms of individual words and bigrams of the previous non-overlapping window for each word are used. Fourier terms are selected for retention by the application of an ANOVA F-test. The retained values are then discretised into words using information gain binning, similar to the MCB step in BOSS. The number of features is further reduced using a chi-squared test after the histograms for each instance are created, removing any words which score below a threshold. It performs a parameter search for $p$ (whether to normalise or not) and over a reduced range of $l$, using a 10-fold cross-validation to determine the performance of each set. The alphabet size  $\alpha$ is fixed to 4 and the $chi$ parameter is fixed to 2. Algorithm~\ref{weasel} gives an overview of WEASEL, although the formation and addition of bigrams is omitted for clarity.


        \begin{algorithm}[h]
    	\caption{WEASEL(A list of $n$ cases of length $m$, ${\bf T}=({\bf X,y})$)}
    	\label{weasel}
    	\begin{algorithmic}[1]
    	\REQUIRE the word length $l$, the alphabet size $\alpha$, the maximal window length $w_{max}$, mean normalisation parameter $p$
    		\STATE Let ${\bf H}$ be the histogram ${\bf h}$
    		\STATE Let ${\bf B}$ be a matrix of $l$ by $\alpha$ breakpoints found by MCB using information gain binning
    		\FOR {$i \leftarrow  1$ to $n$}
    			\FOR {$w \leftarrow  2$ to $w_{max}$}
    				\FOR {$j \leftarrow 1$ to $m-w+1$}
    					\STATE ${\bf o}\leftarrow x_{i,j} \ldots x_{i,j+w-1}$
    					\STATE ${\bf q} \leftarrow$ DFT($o, w, p$) \COMMENT{ {\em {\bf q} is a vector of 	the complex DFT coefficients}}
    					\STATE ${\bf q'} \leftarrow$ ANOVA-F($q, l, y$) \COMMENT{ {\em use only the {\bf l} most discriminative ones}}					
    					\STATE ${\bf r} \leftarrow$ SFAlookup(${\bf q', B}$)
    					\STATE $pos \leftarrow $index(${\bf w, r}$)
    					\STATE ${h}_{i,pos} \leftarrow {h}_{i,pos} + 1$
    				\ENDFOR					
    			\ENDFOR
    		\ENDFOR		
    		\STATE $h \leftarrow \chi^2(h, y)$ \COMMENT{ {\em feature selection using the chi-squared test} }
    		\STATE fitLogistic($h, y$)
    	\end{algorithmic}
        \end{algorithm}
\section{Implementations}

\begin{figure}[!ht]
	\centering
       \includegraphics[width =8cm,clip]{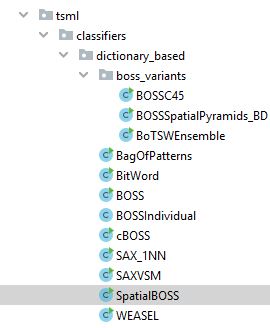}              	
       \caption{Package and class structure for dictionary classifiers in tsml.}
       \label{fig:tsml}
\end{figure}

Dictionary classifiers are available in tsml and sktime, although there is currently a wider range in tsml. Figure~\ref{fig:tsml} shows the class structure for tsml.
All of these inherit from Weka \texttt{AbstractClassifier} and can be used with the standard \texttt{buildClassifier} and \texttt{classifyInstance/distributionForInstance}. Experimental results can be generated and saved in a format consistent with ours using \texttt{Experiments.java}. BagOfPatterns, SAX\_1NN and SAXVSM were all shown to be significantly less accurate than BOSS in the original bake off. S-BOSS is called SpatialBOSS in code. The classifiers in \texttt{boss\_variants} were alternatives evaluated in the BOP to BOSS paper~\cite{large19dictionary}. They were found to be of little use. sktime has a single dictionary classifier in package

\noindent \texttt{classifiers.dictionary\_based} called \texttt{BOSSEnsemble}. This can be configured to behave like BOSS or cBOSS. The equivalence of the BOSS implementations has be demonstrated~\cite{bagnall19toolkits}. S-BOSS and WEASEL are not yet available in sktime, hence the experiments for this part of the redux were all performed in tsml.

\section{Classifier Predictive Performance}
\begin{figure}[!ht]
	\centering
\begin{tabular}{cc}
       \includegraphics[width =5.7cm, trim={3cm 8cm 2.5cm 5cm},clip]{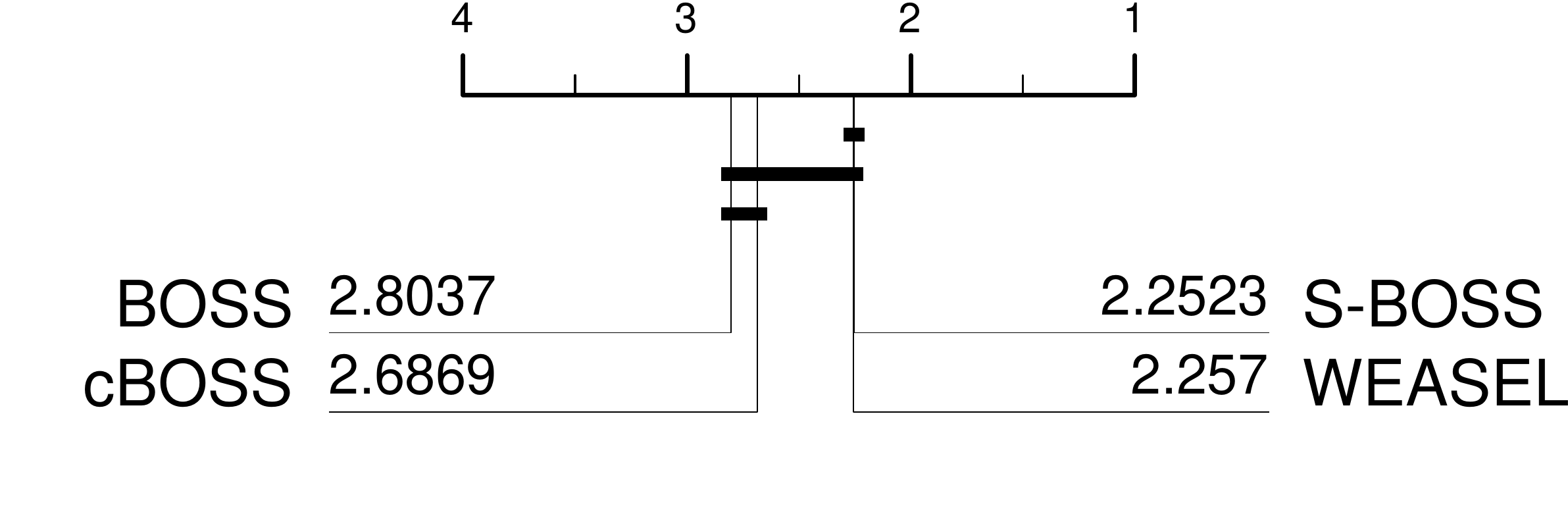}              	
&
       \includegraphics[width =5.7cm, trim={3cm 8cm 2.5cm 5cm},clip]{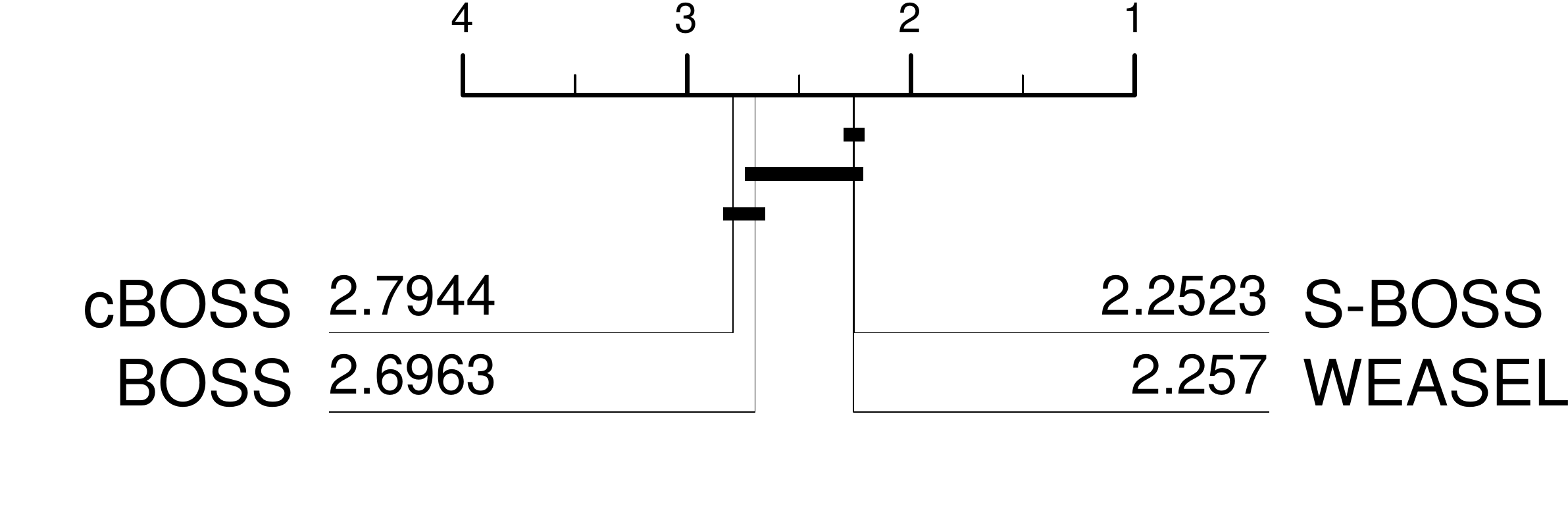}  \\
(a) Accuracy & (b) Balanced Accuracy \\
       \includegraphics[width =5.7cm, trim={3cm 8cm 2.5cm 3cm},clip]{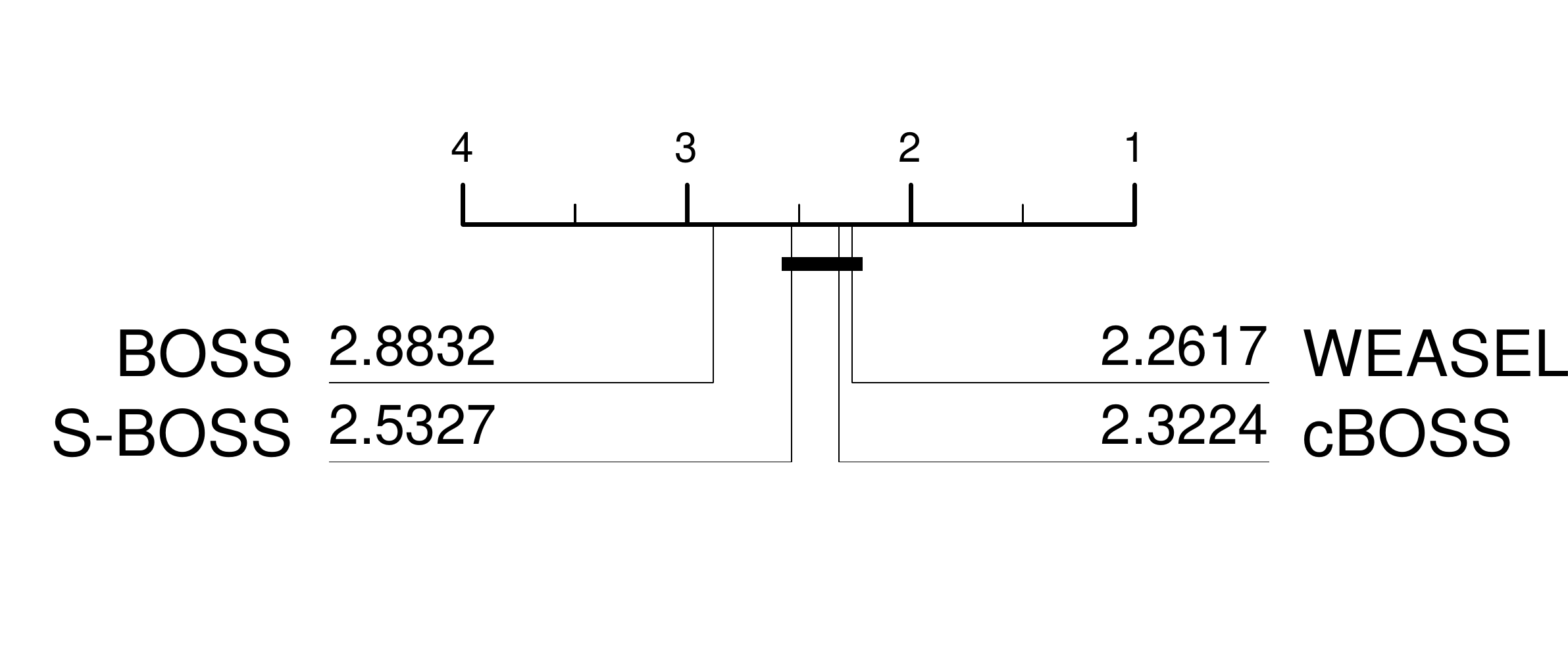}
&
       \includegraphics[width =5.7cm, trim={3cm 8cm 2.5cm 3cm},clip]{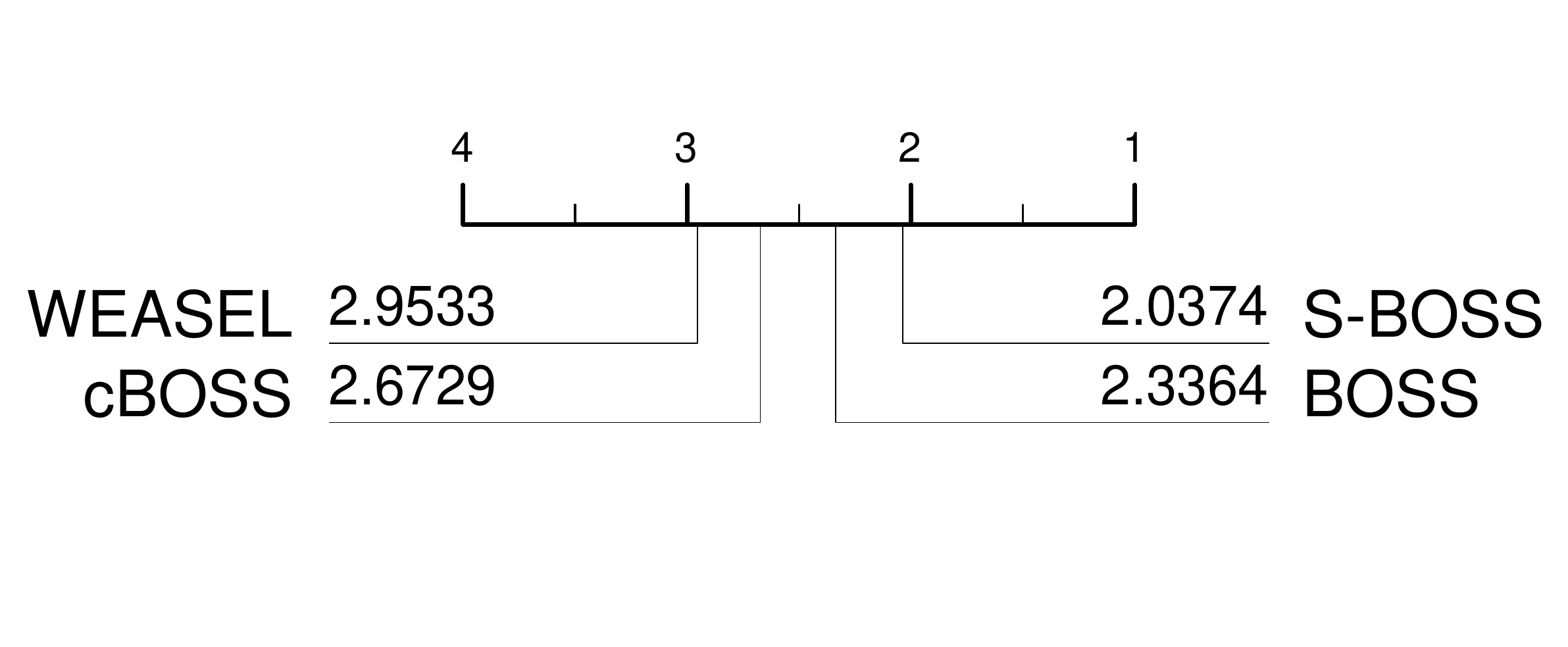}  \\
(c) AUC & (d) NLL \\  
       \end{tabular}
       \caption{Critical difference diagrams for the four dictionary based classifiers on 107 of the 112 equal length/no missing UCR data.}
       \label{fig:prediction}
\end{figure}

Figure~\ref{fig:prediction} shows the accuracy results for four dictionary classifiers: BOSS, cBOSS, S-BOSS and WEASEL. All results are based on the average of 30 resamples of each dataset.  We were only able to complete 107 of the 112 problems within our computational constraints. BOSS and cBOSS completed them all. However, S-BOSS failed to complete FordA, FordB and HandOutlines and WEASEL failed to finish ElectricDevices and NonInvasiveFetalECGThorax2. This immediately highlights usability issues with both of these algorithms for large data, which we address in more detail in Section~\ref{sec:timeandspace}. We compare classifiers on unseen data based on the quality of the decision rule (using classification error and balanced classification error to account for class imbalance), the ability to rank cases (with the area under the receiver operator characteristic curve) and the probability estimates (using negative log likelihood). Summaries of the complete results are available from the accompanying website\footnote{http://timeseriesclassification.com/results/bakeoff-redux/dictionary-results.zip}.

In terms of accuracy, Figure~\ref{fig:prediction} demonstrates that there is little difference between BOSS and cBOSS nor between S-BOSS and WEASEL. The solid lines represent classifiers within which there is no significant difference. This makes sense if these diagrams are used as Demsar initially proposed~\cite{demsar06comparisons}, since the cliques are formed through a global post-hoc test. However, this post-hoc test is dependent on the number of classifiers, and is not thought to be the best approach~\cite{benavoli16pairwise}. Instead, we perform a pairwise test using a Holm correction. This can (and does) give rise to the occasional anomaly. Suppose we have three classifiers $A$, $B$ and $C$ with average rank $A>B>C$. It is possible that in a pairwise test $A$ is not significantly different to $C$, but that $B$ is significantly different. This is what has happened with the WEASEL, cBOSS, BOSS clique in Figure~\ref{fig:prediction} (a).  Table 2 shows the decision matrix for the signed Wilcoxon rank test prior to the correction for multiple tests to demonstrate this.


\begin{table}[!ht]
\label{tab:decisions}
\caption{Wilcoxon decisions for pairwise test, where true indicates significant difference and the p-value (prior to correction) is shown in brackets.}
	\centering
\begin{tabular}{c|cccc}
            & S-BOSS    &  WEASEL           & cBOSS         & BOSS \\ \hline
S-BOSS      & 	        &   false (0.463)	& true (0.005) & true (0.0005) \\
WEASEL      & false	    & 	                & true	(0.0245) & false (0.048) \\
cBOSS       & true	    &   true	        &               & false (0.877)  \\
BOSS        & true	    &   false	        &   false        &       \\ \hline
\end{tabular}  
\end{table}

Critical difference diagram is a misnomer for these plots, since we are not forming cliques by a critical difference. However, we believe there it is still helpful displaying these graphs, as they give a clear indication of relative performance. The general conclusion is that there is no overall difference in accuracy between S-BOSS and WEASEL, both of which are better than cBOSS and BOSS, which also form a clear group.

The cliques formed on these graphs are not the end point of the analysis of performance. Ultimately, we want to understand the differences between the classifiers, not just choose between them. We come back to the accuracy results later, but first the other three CD diagrams contain results worthy of note. The balanced accuracy results very closely mirror the accuracy diagram. There is no evidence of different performance for problems with class imbalance. The AUC results indicate no difference in performance between S-BOSS, cBOSS and WEASEL. Why is cBOSS relatively better at ranking cases than it is at predicting cases? It may be that differences are harder to detect with AUC than accuracy, or that although cBOSS ranks the data well, it is poor at classifying around the decision boundary.

The NLL results are also surprising. WEASEL is the worst algorithm at producing probability estimates, despite its accuracy. WEASEL tends to produce extreme probabilities, and if the classification is wrong, this invokes a large NLL penalty. This would be a particular problem for an algorithm such as HIVE-COTE~\cite{lines18hive} which ensembles probability estimates. Similarly, cBOSS performs well when we measure AUC but poorly when NLL are compared. This difference could help identify areas for algorithmic improvement.

\begin{figure}[!ht]
	\centering
\begin{tabular}{cc}  
       \includegraphics[width =6cm,trim={2cm 0cm 2cm 0cm},clip]{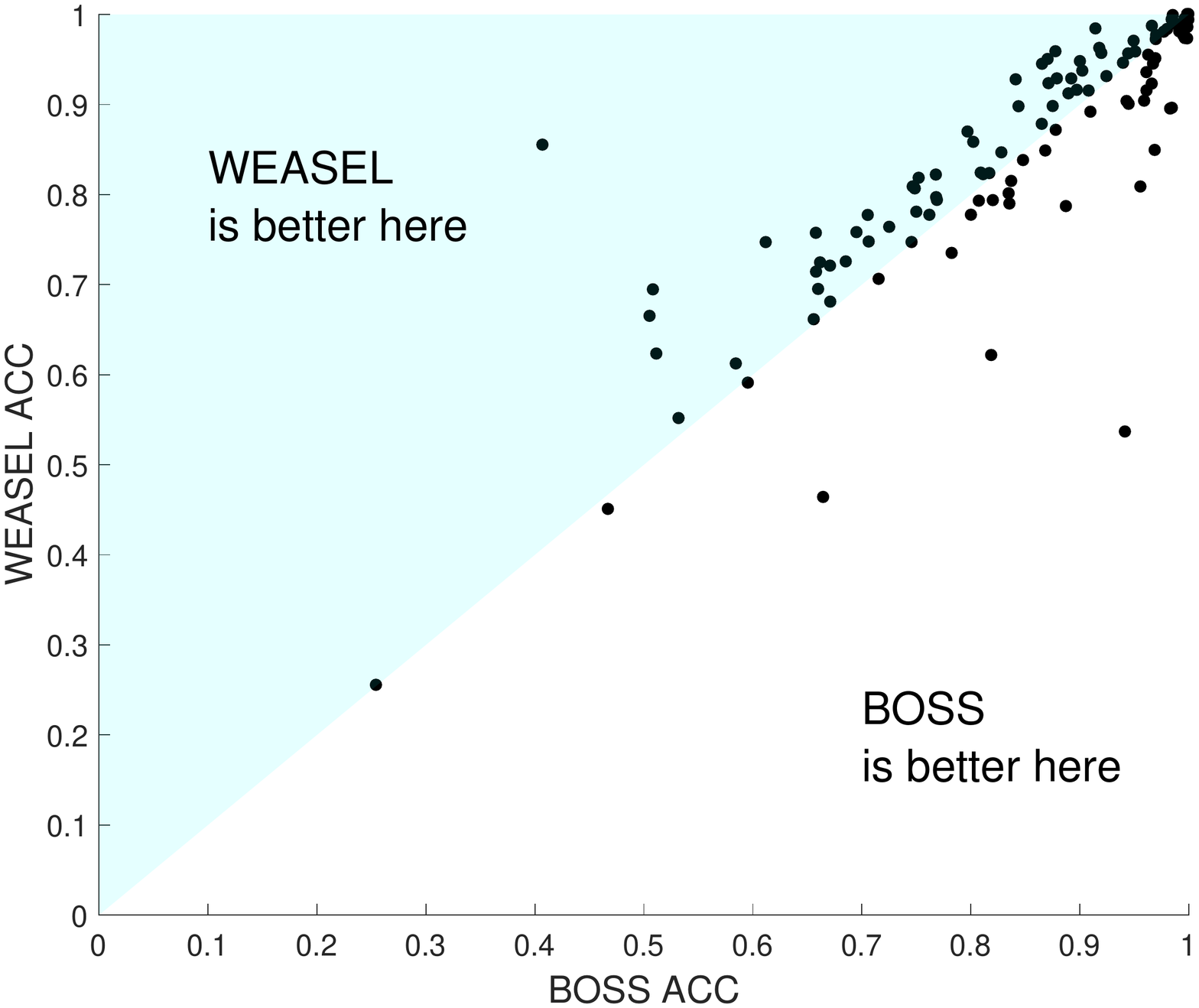}              	
&
       \includegraphics[width =6cm, trim={2cm 0cm 2cm 0cm},clip]{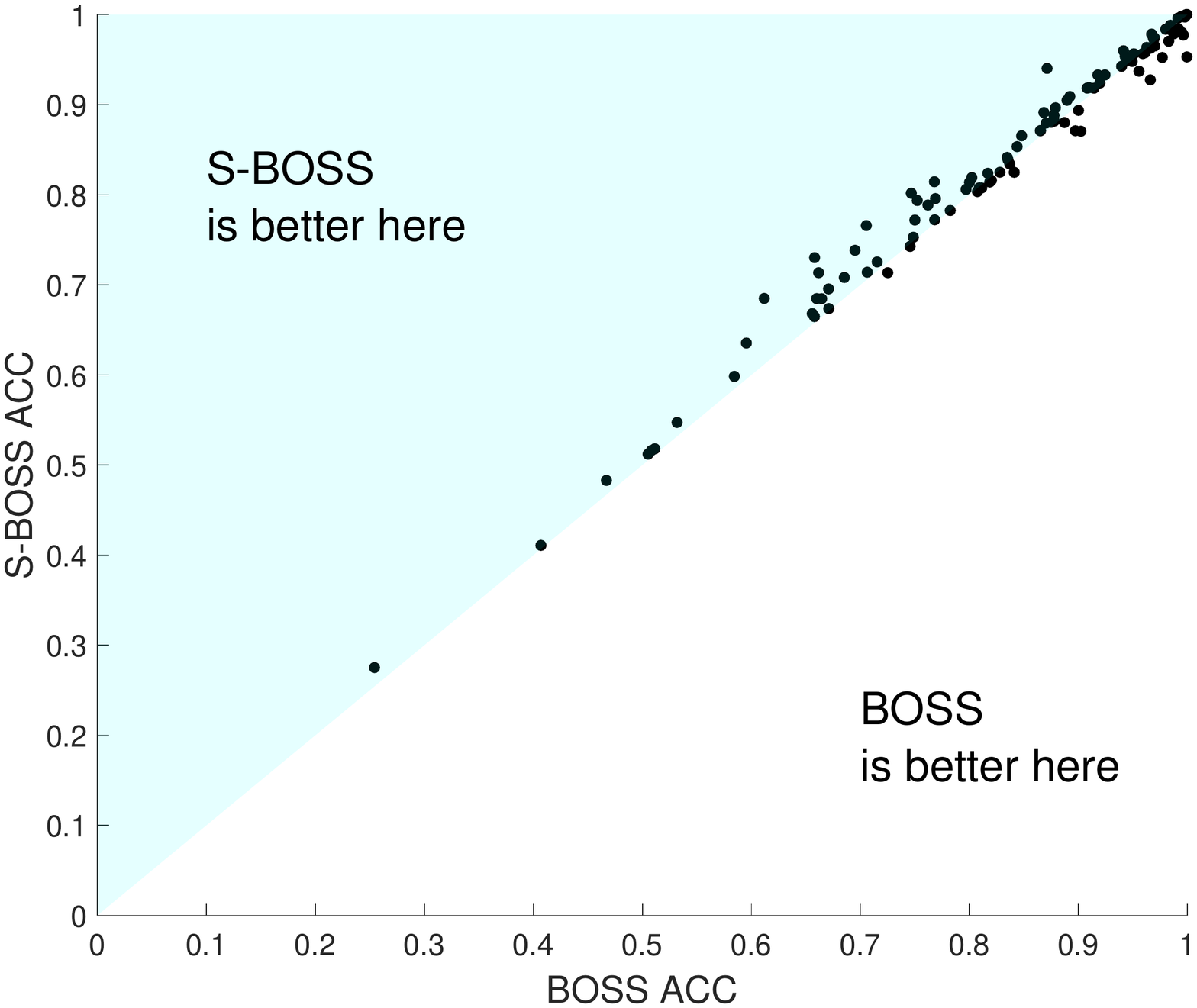} \\
 W/D/L: 67/1/42 &    W/D/L: 74/0/36
       \end{tabular}
       \caption{Scatter plots of S-BOSS vs WEASEL and BOSS vs cBOSS, and the win/draw/loss (W/D/L) of the top classifier over the bottom classifier.}
       \label{fig:scatter_dict1}
\end{figure}

\begin{figure}[!ht]
	\centering
	\centering
\begin{tabular}{cc}  
       \includegraphics[width =6cm,trim={2cm 0cm 2cm 0cm},clip]{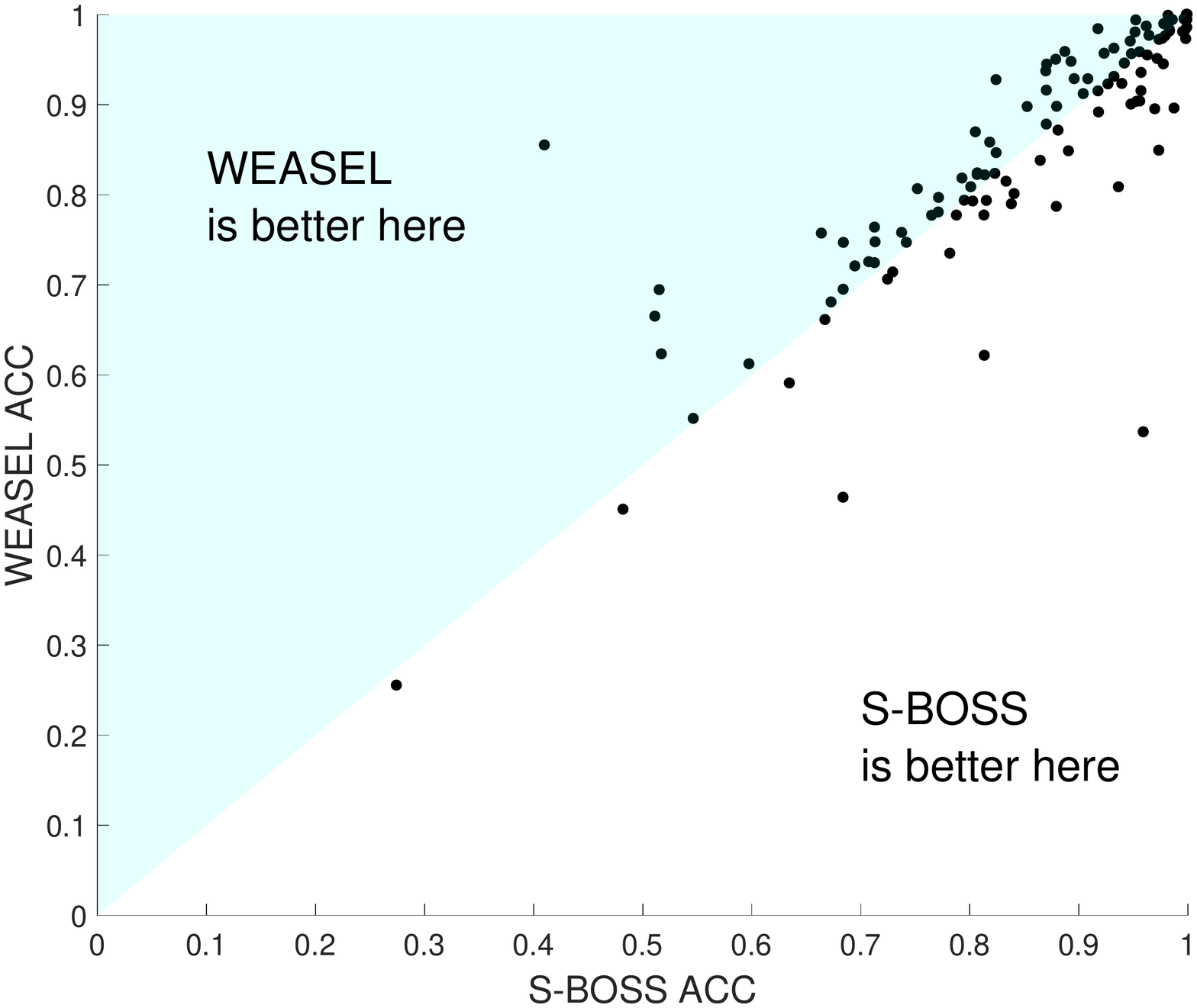}              &
       \includegraphics[width =6cm, trim={2cm 0cm 2cm 0cm},clip]{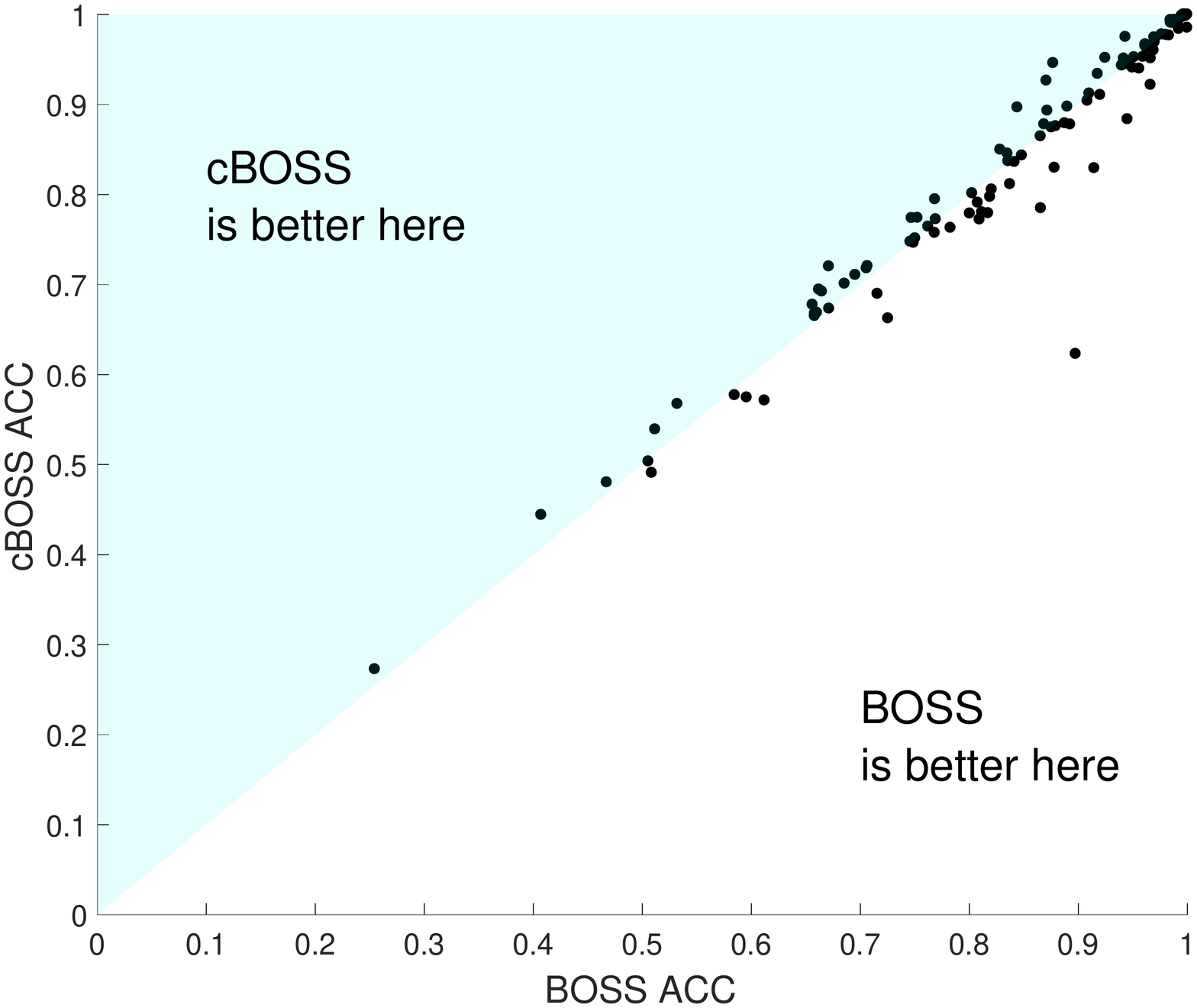} \\
W/D/L: 52/0/58 & W/D/L: 50/2/56
\end{tabular}
       \caption{Scatter plots of S-BOSS vs WEASEL and BOSS vs cBOSS, and the win/draw/loss (W/D/L) of the top classifier over the bottom classifier.}
       \label{fig:scatter_dict2}
\end{figure}

Differences in algorithms are best highlighted through a more detailed analysis.  Figure~\ref{fig:scatter_dict1} shows the scatter plots for S-BOSS and WEASEL against BOSS. The patterns are quite different. S-BOSS consistently improves BOSS by a small margin, whereas the WEASEL-BOSS difference has a high variance. WEASEL seems to use different underlying features to classify. S-BOSS is much more like BOSS, but it can mitigate against areas of the series confounding the classifier by isolating different regions in time. Figure~\ref{fig:scatter_dict2} shows the plots for S-BOSS vs WEASEL and cBOSS vs BOSS. The former demonstrates the fundamental underlying difference between S-BOSS vs WEASEL whereas the latter confirms that cBOSS is very like BOSS, except it is much faster (as shown in Section~\ref{sec:timeandspace}).

\section{Classifier Time and Space Complexity}
\label{sec:timeandspace}

Both S-BOSS and WEASEL are better than BOSS over a range of problems, but at what cost in terms of memory and run time? To test this we use a data simulator to generate randomised problems that are designed to be optimal for a dictionary based approach. We select two different shapes, then  place the shapes in each series at random locations. The frequency of each shape is different in each class. We embed the shapes in random noise. We can then measure the run time and space complexity for varying train set size, test set size and series length. We also record accuracy, but we note that we have constructed the problem so that standard BOSS should be the best approach to solving it. Hence, we are not expecting anything to be more accurate than BOSS unless it is some kind of edge case.

\begin{figure}[!ht]
	\centering
\begin{tabular}{cc} 
       \includegraphics[width =6cm,trim={2cm 8cm 2cm 9.5cm},clip]{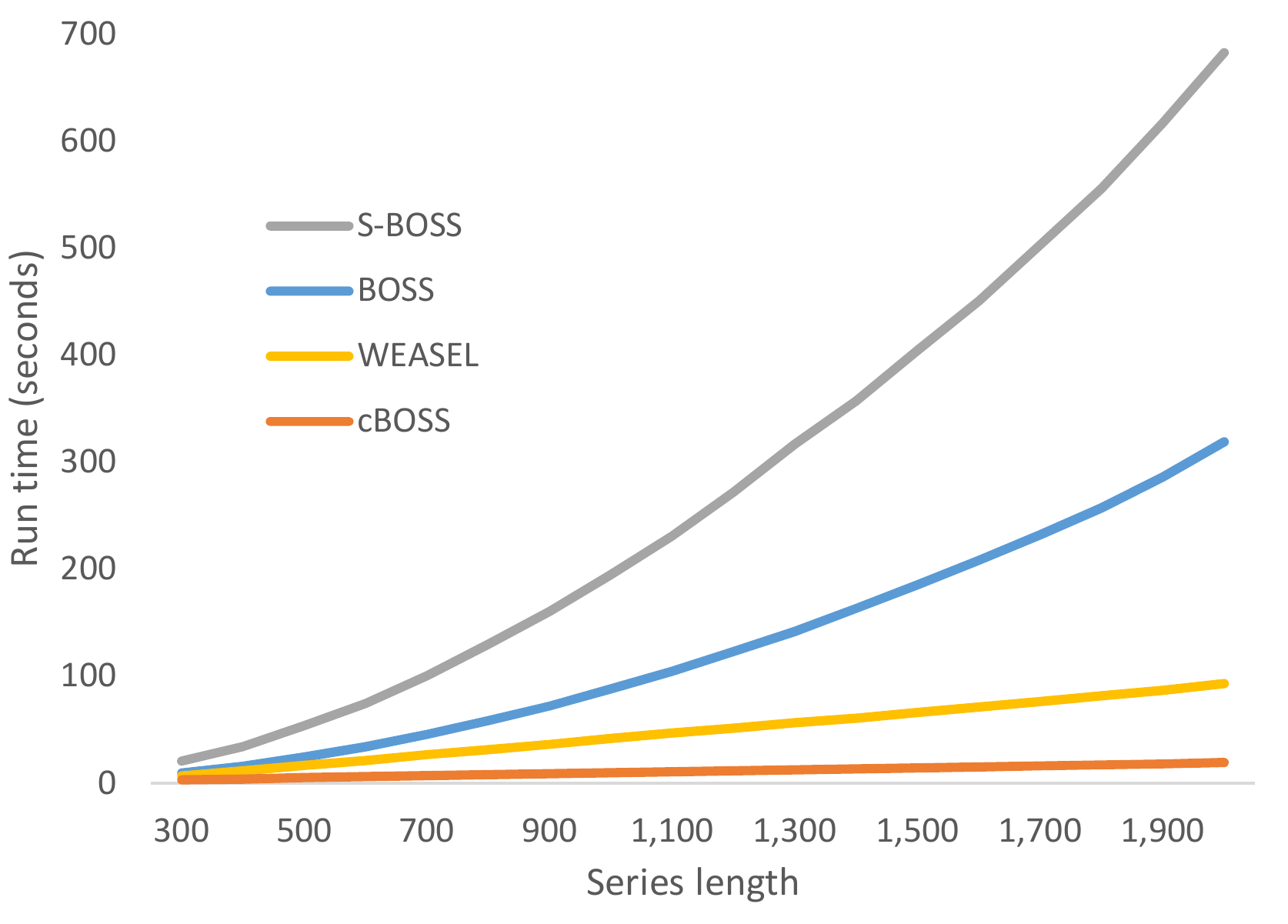}              	
&
       \includegraphics[width =6cm, trim={2cm 8cm 2cm 9.5cm},clip]{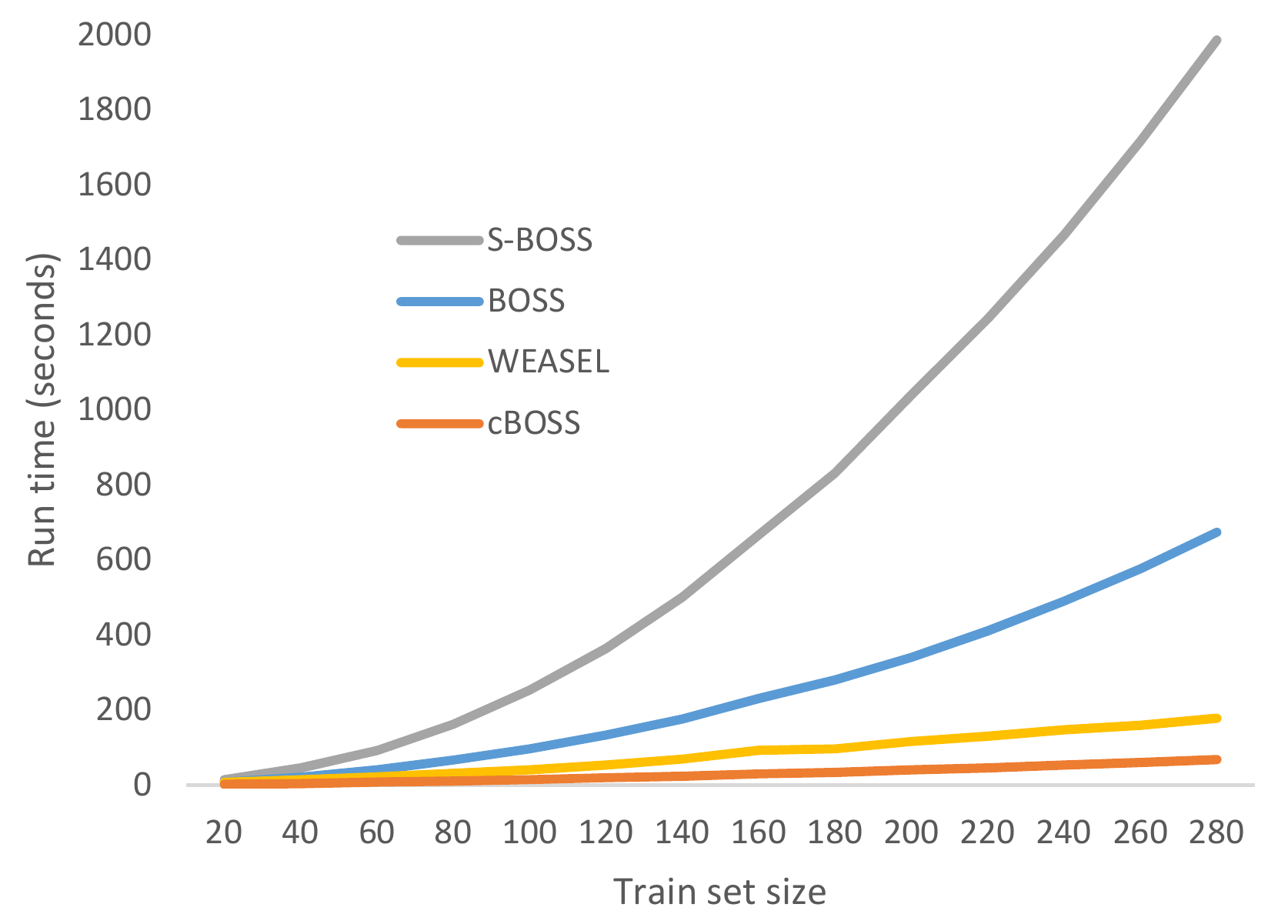}
       \end{tabular}
       \caption{Run time for four dictionary classifiers for changing series length and train set size on simulated data.}
       \label{fig:dict_times}
\end{figure}

Figure~\ref{fig:dict_times} show the run time for the four dictionary classifiers for changing train set size and series length.

S-BOSS is the slowest algorithm. This is not unexpected given that it does all the work of the BOSS classifiers, and then further work to also end up with longer feature vectors. The run time complexity of both BOSS and S-BOSS is quadratic for series length for fixed number of cases and vice versa. However, S-BOSS takes approximately twice as long as BOSS for any fixed series length and train set size. cBOSS is approximately ten times faster than BOSS and is the fastest algorithm. WEASEL is about twice as fast.

\begin{figure}[!ht]
	\centering
\begin{tabular}{cc} 
       \includegraphics[width =5.7cm,trim={2cm 8cm 2cm 9cm},clip]{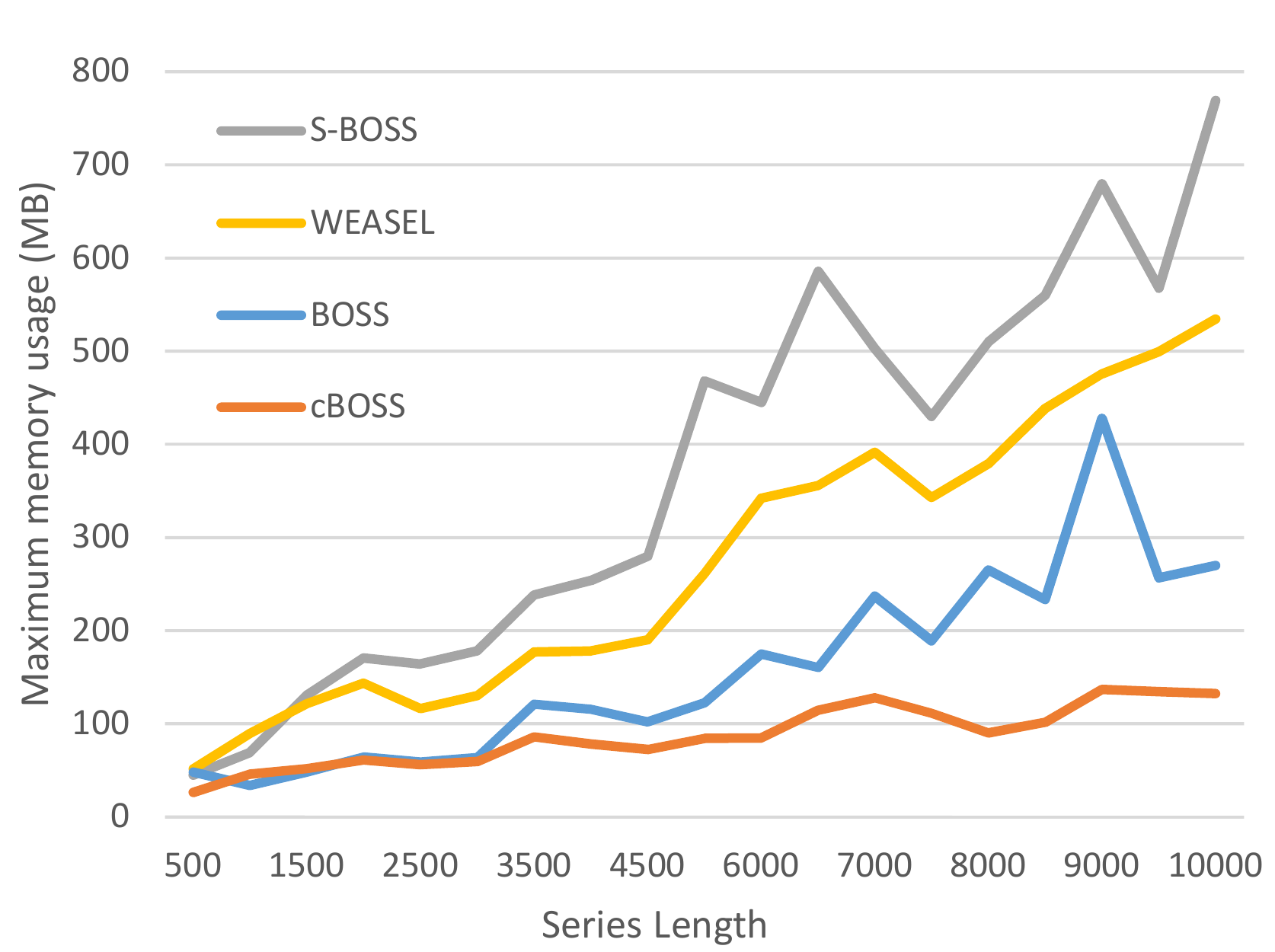}              	
&
       \includegraphics[width =5.7cm, trim={2cm 8cm 2cm 9cm},clip]{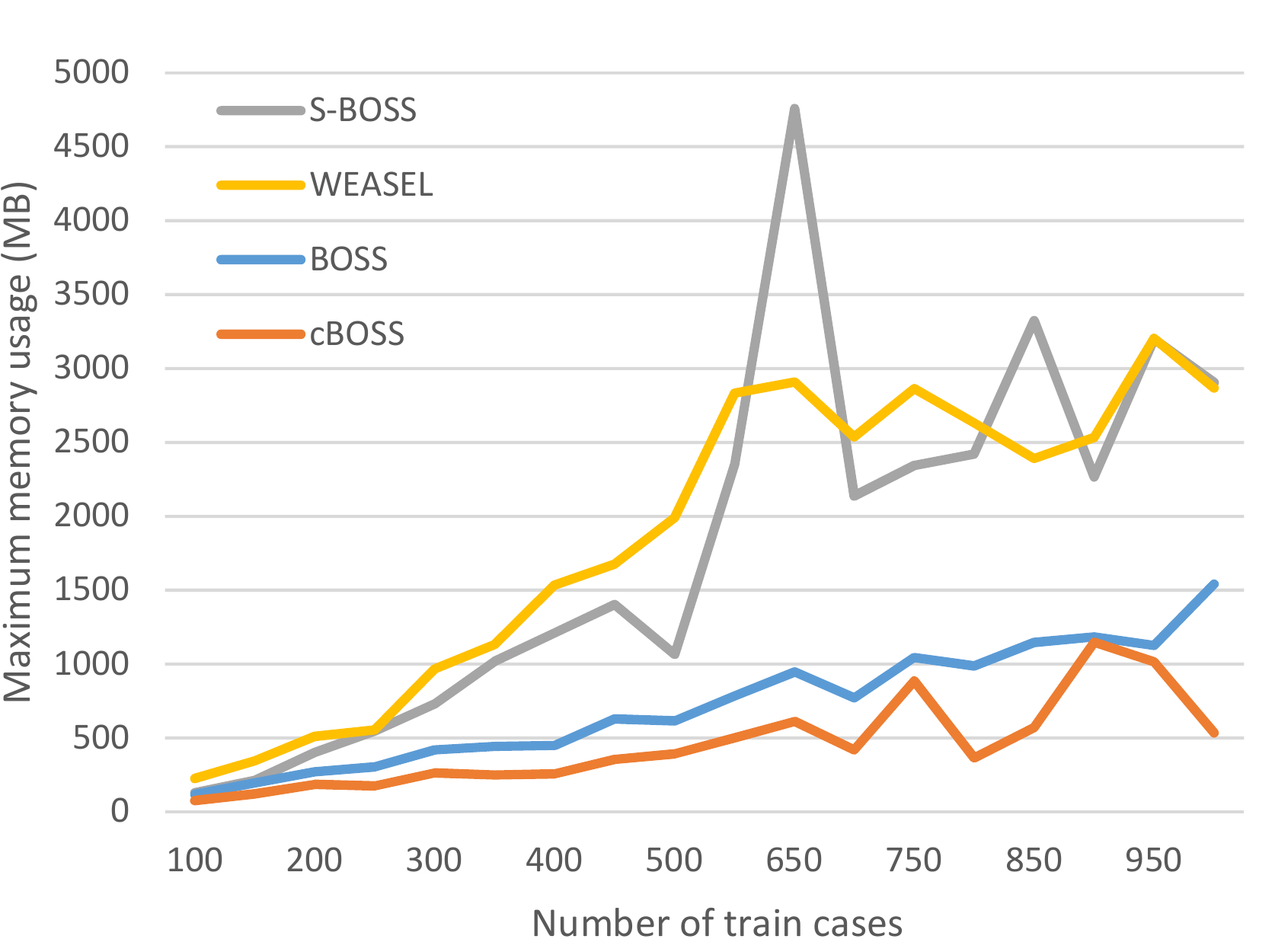} \\
       (a) & (b)
       \end{tabular}
       \caption{Memory for the four dictionary classifiers for changing series length (a) and train set size (b) on simulated data.}
              \label{fig:dict_mem}
\end{figure}
Figure~\ref{fig:dict_mem} shows the memory requirements for the four algorithms with changing series length and train set size. These are estimates of the maximum memory used as measured by notifications from the garbage collector (see class \texttt{MemoryManager.java}). The nature of these algorithms means that the peak memory is usually much higher than the final memory. cBOSS uses the least memory. BOSS on average uses about double the memory of cBOSS. WEASEL and S-BOSS need about four times the memory of cBOSS. S-BOSS is more sensitive to series length than WEASEL, due to the fact longer series are likely to create more partitions.


\section{Conclusions}
\label{sec:concs}
Although S-BOSS is the most accurate approach, it is also the slowest, and requires the most memory. WEASEL performs as well as S-BOSS on average, although it produces diverse results when compared to the other algorithms in the same class. WEASEL is faster than S-BOSS and BOSS, but equally memory intensive as S-BOSS. For problems with a large number of cases or very long series, cBOSS may be the best choice, particularly as it is possible to contract the run time prior to running the algorithm. This option allows for checkpointing (stopping and restarting) in order to control the execution time.

These classifiers are all memory intensive. For BOSS, S-BOSS and WEASEL this is primarily a problem of the maximum memory required during the search. It seems likely that this could be optimised to reduce the required working memory, particularly for WEASEL in the feature selection stage. It also seems sensible to apply the Spatial pyramid approach to cBOSS, to see if we can get the improved accuracy for a lower memory overhead. At the heart of the memory problem for the BOSS variants is the requirement to store a version of the data for each ensemble member in order to classify with the nearest neighbour classifier. Previous experiments investigating changing 1-NN for a model based classifier were unsuccessful~\cite{large19dictionary}. However, given our improved understanding of how these classifiers work, it is worth investigating this aspect again.

We end on a cliff hanger. BOSS is one of five core classifiers in HIVE-COTE~\cite{lines18hive}. An obvious question is what effect is there on HIVE-COTE from changing from BOSS to cBOSS, S-BOSS or WEASEL. Can we get a significant improvement to HIVE-COTE? Watch this space ...

\bibliographystyle{plain}

\end{document}